\documentclass{article}

\usepackage{arxivEN}

\usepackage[utf8]{inputenc}
\usepackage[T1]{fontenc}

\usepackage{graphicx}

\usepackage[
	backend=biber,
	style=authoryear,
	citestyle=authoryear,
	hyperref
]{biblatex}
\usepackage{csquotes}
\usepackage[english]{babel}

\usepackage{booktabs}
\usepackage{siunitx}

\usepackage{hyperref}
\usepackage{url}

\graphicspath{{figures/}, {images/}}

\title{%
    Multimodal Semantic Transfer\\
    from Text to Image.\\
    Fine-Grained Image Classification\\
    by Distributional Semantics.}

\author{
    Simon Donig \href{https://orcid.org/0000-0002-1741-466X}{\includegraphics[scale=0.06]{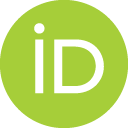}}\\
    Chair for Digital Humanities\\
    University Passau, Germany\\
    \url{simon.donig@uni-passau.de}\\
    \And
    Maria Christoforaki \\
    Chair for Data Science\\
    Institute for Computer Science\\
    University of St.Gallen, Switzerland \\
    \url{maria.christoforaki@unisg.ch} \\
    \AND
    Bernhard Bermeitinger \href{https://orcid.org/0000-0002-2524-1850}{\includegraphics[scale=0.06]{orcid.png}} \\
    Chair for Data Science\\
    Institute for Computer Science\\
    University of St.Gallen, Switzerland \\
    \url{bernhard.bermeitinger@unisg.ch} \\
    \And
    Siegfried Handschuh \\
    Chair for Data Science\\
    Institute for Computer Science\\
    University of St.Gallen, Switzerland \\
    \url{siegfried.handschuh@unisg.ch} \\
}

\date{December 2019}

\hypersetup{
	pdftitle={Multimodal Semantic Transfer from Text to Image. Fine-Grained Image Classification by Distributional Semantics.},
	pdfauthor={Simon Donig, Maria Christoforaki, Bernhard Bermeitinger, Siegfried Handschuh},
	pdfkeywords={deep learning, artificial neural networks, digital humanities, distributional semantics, neoclassicism}
}

\begin{document}

\maketitle

\section{Introduction}
In the last years, image classification processes like neural networks in the area of art-history and \emph{Heritage Informatics} have experienced a broad distribution \parencite{lang_AttestingSimilaritySupportingOrganizationStudy_2018}.
These methods face several challenges, including the handling of comparatively small amounts of data as well as high-dimensional data in the Digital Humanities. In most cases, these methods map the classification task to flat target space. This \enquote{flat} surface loses several relevant dimensions in the search for ontological uniqueness, including taxonomical, mereological, and associative relationships between classes, or the non-formal context, respectively.

The proposed solution by \citeauthor{donig_VomBildTextUndWieder_2019a} \parencite{donig_VomBildTextUndWieder_2019a} to expand the capabilities of visual classifiers is to take advantage of the greater expressiveness of text-based models. Here, a \emph{Convolutional Neural Network} (CNN) is used that output is not as usual a series of flat text labels but a series of semantically loaded vectors. These vectors result from a \emph{Distributional Semantic Model} (DSM) (\cite{lenci_DistributionalModelsWordMeaning_2018a}) which is generated from an in-domain text corpus.

Here, we propose an early implementation of this method and analyze the results.

The conducted experiment is based on the collation of two corpora: one text-based and a visual. From the text, a DSM is created and then queried for a list of target words that are functionally the labels that are manually given to the images. The result is a list of vectors that correspond to the target words leading to images that are annotated not only with a label but also with a unique vector. The images and vectors are used as training data for a CNN that, afterward, should be able to predict a vector for an unseen image. This prediction vector can be converted back to a word by the DSM using a nearest-neighbor algorithm. We are looking for richer representation in this process, so we choose the five nearest neighbors. The similarity measure is the cosine similarity for high-dimensional vector spaces between the given target vector and the prediction vector. We derive a positive classification result if the target label is within the list of five nearest neighbors of the prediction vector.

Moreover, we compare the results between the proposed classification method and a conventional classification method using the same CNN as for the vector-based experiment but a list of flat labels. Finally, we can show that the vector-based approach (judging from classification metrics) is equally performant or even better than the label-based version.

\section{Experiment Structure}
The experiment is based on one text and one visual corpus from the area of material culture research with a focus on neo-classical artifacts.

\subsection{Text Corpus}
The text corpus is compiled out of 44 sources that are available under a free and permissive license. It contains English specialist publications on furniture and spatial art, published from the end of the 19\textsuperscript{th} century to the middle of the 20\textsuperscript{th} century. In multiple steps, the corpus is cleaned and preprocessed. First, a series of standard NLP methods are applied like tokenization, sentence- and word splitting, normalization of numbers, and named entity recognition (NER). Since we used retro-digitized material from a different source, we implemented manual corrections for the most common errors (such as ligatures like \texttt{TT} that were misinterpreted as \texttt{U}). Another level of preprocessing consists of content-related augmentations. In particular, we normalized compound words and synonyms according to a specified list, which is based on an ontology, the \emph{Neoclassica-Ontology} \parencite{donig_NeoclassicaMultilingualDomainOntology_2016}. This resulted in the final text corpus of total \num{3067237} words comprised of \num{107518} basic word forms.

The DSM is created using the \emph{Indra Frameworks} \parencite{sales_IndraWordEmbeddingSemanticRelatedness_2018a} with a vector size of \num{50}, a word window size of \num{10}, and minimal word count of \num{5}. We used \emph{Skipgram} as the \emph{Word2Vec} model \parencite{mikolov_EfficientEstimationWordRepresentationsVector_2013} with negative sampling.

\subsection{Image Corpus}
The image corpus consists of \num{1231} images of neoclassical furniture in their entirety, which are permissive licensed\footnote{The corpus was compiled from collections from the Metropolitan Museum, New York, the Victoria \& Albert Museum, London, the Wallace Collection, London, and several contemporary pattern books.}. The images are either historical pictorial material or photographs from the modern as-built documentation. They are split into \num{28} different classes.

\subsection{Combined Corpus}
The nature of the experiment is \emph{proof-of-concept}, so we used a VGG-like architecture of a \enquote{simple} convolutional neural network\footnote{The CNN is built from three convolutional blocks with two consecutive Convolutional Layers each with 32/64/64 filters of size $ 3 \times 3 $. Each block is followed by a Maximum Pooling Layer with a size of $ 2 \times 2 $ and a Dropout Layer for regularization with a dropout rate of \num{0.25}. After the Convolutional Block, there are two Fully Connected Layers with \num{256} nodes and a Dropout Layer with a dropout rate of \num{0.5}. The weights and biases for the Convolutional and the Fully Connected Layers are initialized randomly. Their activation function is \emph{ReLU}. The last layer, the classification layer, is a Fully Connected Layer with \num{50} output nodes and a linear activation function. The optimization function of the chosen \emph{mean absolute error} function is \emph{RMSprop}. The implementation is done with Keras (\url{https://keras.io}) and TensorFlow (\url{https://tensorflow.org})}.

The independence and robustness of the train/test split are guaranteed with 5-fold cross-validation on \SI{80}{\percent} on the full corpus from which \SI{80}{\percent} are used for training and \SI{20}{\percent} for testing. The remaining \SI{20}{\percent} are treated as an evaluation set.

\begin{figure}
    \centering
    \includegraphics[height=.45\textheight]{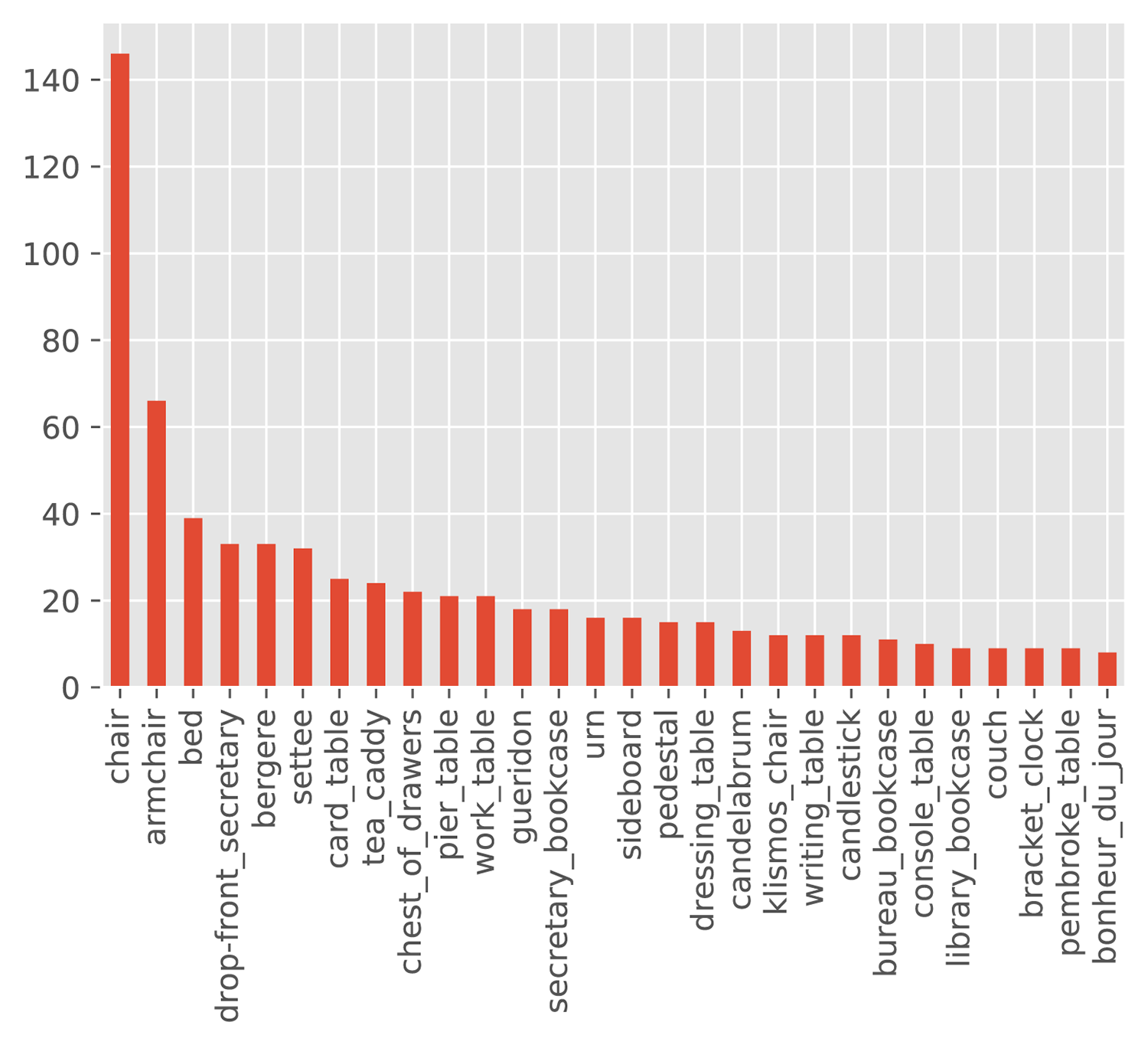}
    \caption{Class distribution of the image corpus}
    \label{fig:class_distribution}
\end{figure}

The dataset is unbalanced, as can be seen in Fig.~\ref{fig:class_distribution}. During training, more prominent classes are weighted down and underrepresented classes are given a higher weight \parencite[p.~27]{johnson_SurveyDeepLearningClassImbalance_2019}. Apart from dropout during training for regularization, \emph{Early Stopping} was used to prevent overfitting.

\section{Results}

\subsection{Results for CNN trained on Word Vectors}

The Top-5 true-positive rate is \num{0.6052}, meaning that the gold label from the annotations is in \SI{60}{\percent} of the cases within the list of the five nearest neighbors.

However, the mathematical quality metric in itself represents only part of the overall picture. For this reason, a qualitative analysis of the results was performed.

A few true-positives show that, for example, the classification is by no means random but that the top 5 terms originate from the same semantic neighborhoods. They express several relationships of a taxonomical and associative nature.

\begin{figure}
    \centering
    \includegraphics[height=.4\textheight]{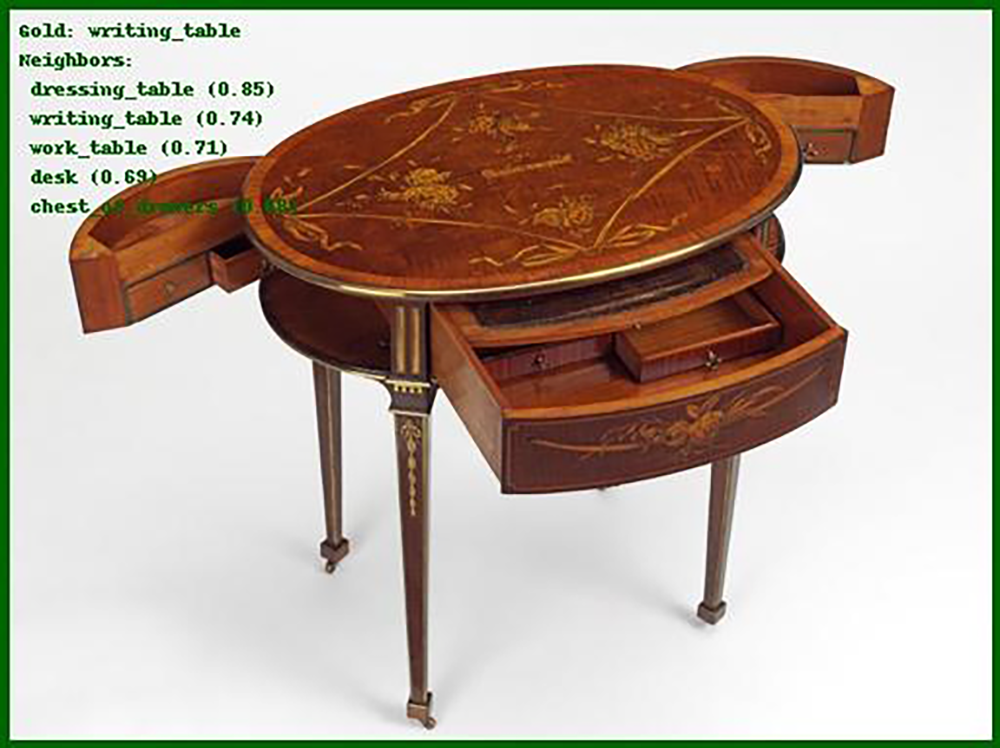}
    \caption{Differences in the classification of the same object (1)}
    \label{fig:classification_same_object_1}
\end{figure}

\begin{figure}
    \centering
    \includegraphics[height=.4\textheight]{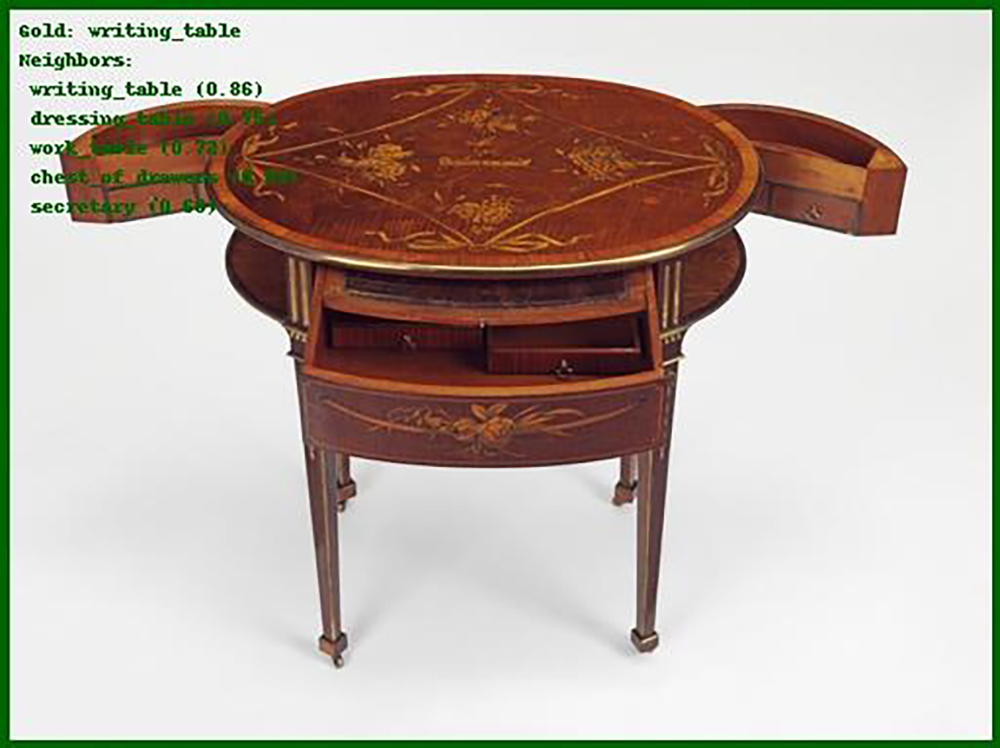}
    \caption{Differences in the classification of the same object (2)}
    \label{fig:classification_same_object_2}
\end{figure}

As an example, the \emph{Roentgen} desk from the \emph{Victoria \& Albert} inventory in Fig.~\ref{fig:classification_same_object_1} is associated with the labels \texttt{dressing\_table}, \texttt{writing\_table} and \texttt{work\_table}. This triad is meaningful because many of these artifacts were multifunctional and fulfilled several of these functions. Besides, those artifacts that decidedly served only one purpose are constructively similar to the other types of furniture. The similarity of the three concepts thus emerges both on a semantic level (the distance of the words in the DSM, which in turn is the product of real-world distance) and on a visual level in the CNN (visual similarity of the form). Another image of the same object (see Fig.~\ref{fig:classification_same_object_2}) shows, on the other hand, that the method is consisted in itself---the top 4 nearest neighbors of the predicted vector are identical, although the photograph is taken from a different perspective---and, on the other hand, that the visual features within the CNN also affect the classification process. Since writing cabinets (\emph{secrétaires à abbatants}) are often displayed frontally, upright and with an open flap or drawer, their appearance in the image seems to have triggered a classification as a secretary. In the first image, however, the presence of drawers could have led to a classification as a chest-of-drawers, which is associated with drawers on a semantic level.

\begin{figure}
    \centering
    \includegraphics[height=.6\textheight]{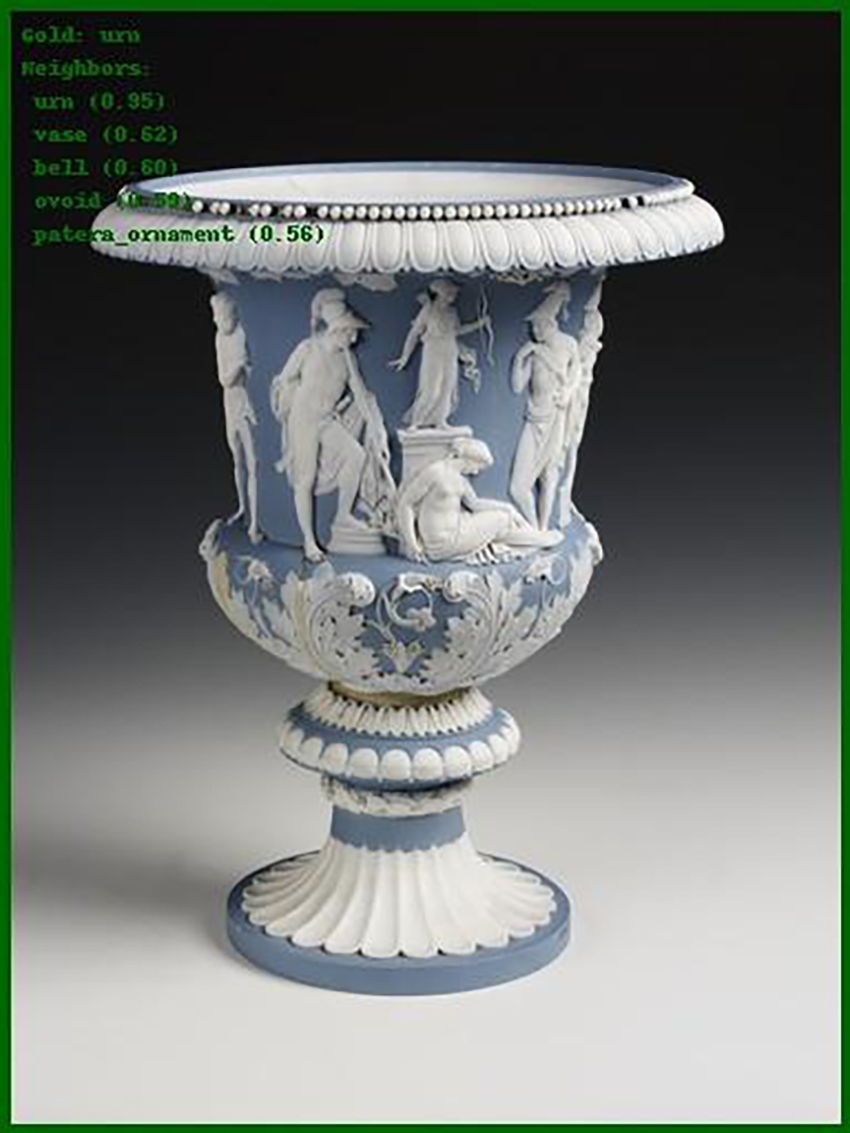}
    \caption{A sèvres copy of a Medici vase results in a classification of associative labels.}
    \label{fig:medici_vase}
\end{figure}

While the labels considered so far reflect the taxonomic relationships and all originate from the target words derived from the ontology, Fig.~\ref{fig:medici_vase} shows that the procedure can also generate labels for itself, purely data-driven. The crater vase shown was originally classified as an urn. The Top-2 words therefore also reflect taxonomic relationships (\enquote{urn}, \enquote{vase}). The other concepts reflect associative relationships. The label \texttt{bell} is a leftover from the data cleaning process of the text corpus to describe this kind of artifact. \enquote{Ovoid} refers to the egg stick decoration of the upper bead, which is often described with this adjective. This ornamentation seems at the same time to have an association with the rosette (patera). In this way, the target word \texttt{patera\_element} appears among the top-5 nearest neighbors, although only whole artifacts were annotated in the image corpus but not their decoration.

\begin{figure}
    \centering
    \includegraphics[height=.6\textheight]{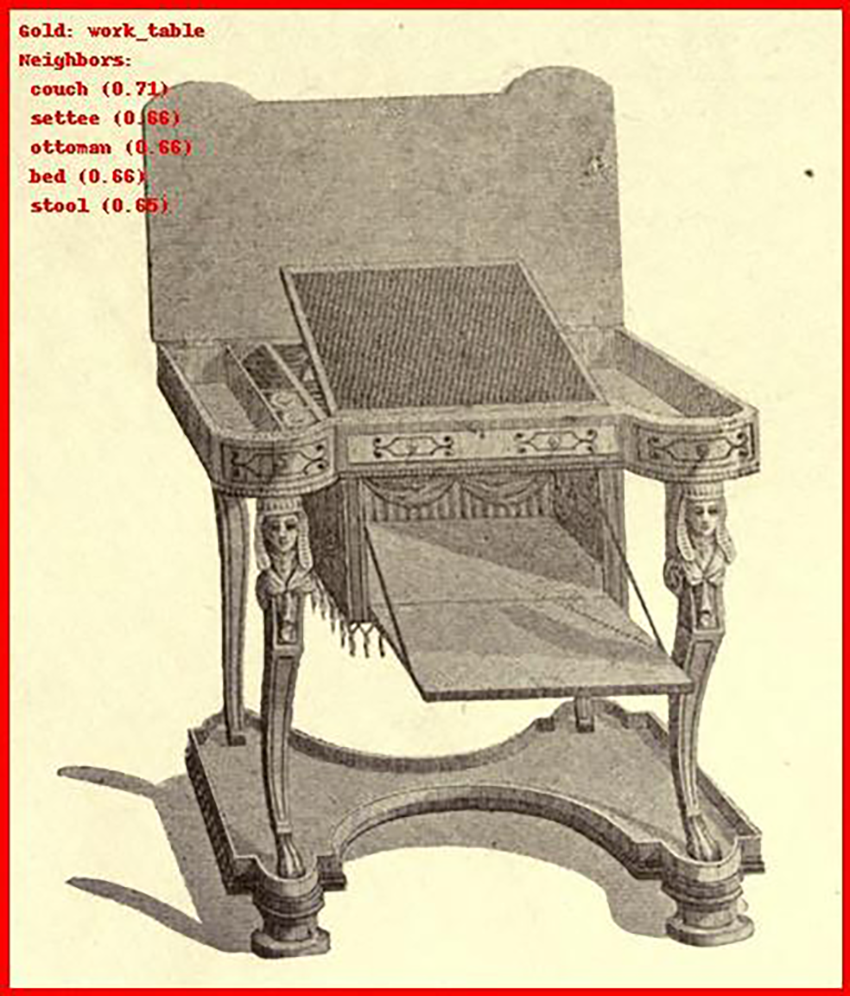}
    \caption{Misclassification of a sewing table in the proximity of seating furniture hierarchy.}
    \label{fig:misclassification}
\end{figure}

An effect of the visual classifier cannot be excluded, as shown in Fig.~\ref{fig:misclassification}. The misclassification of the object, a small sewing table, led to consistent attributions in the area of seating and reclining furniture. Looking at the outer form of the artifact on a more abstract level, visual proximity to e.g. a (double)-camel-back sofa can be easily derived.

\subsection{Comparison to a CNN with flat labels}

To better estimate the differences between the two approaches (conventional classification by flat labels vs.~vector-based approach), we present a comparison between different metrics for both approaches.

\begin{table}
    \centering
    \begin{tabular}{lrr}
        \toprule
        Metric                     &   Vector-based &   Label-based \\ \midrule
        Top-1 Accuracy             &   \num{0.5378} &  \num{0.5302} \\
        Top-1 Precision            &   \num{0.3487} &  \num{0.4446} \\
        Top-1 Recall               &   \num{0.2974} &  \num{0.4044} \\
        Top-1 F1-Score             &   \num{0.2976} &  \num{0.3985} \\ \midrule
        Top-5 True-Positive-Rate   &   \num{0.6052} &  \num{0.8463} \\
        Top-5 False-Positive-Rate  &   \num{0.3948} &  \num{0.1537} \\
         \bottomrule
    \end{tabular}
    \caption{Results of different Metrics.}
    \label{tab:results}
\end{table}

As shown in Table~\ref{tab:results} the metrics are comparable.

Thus, the proposed approach not only improves the accuracy but also provides a richer description of the image.

\section{Conclusion and Outlook}
In this paper, we have presented a new multimodal approach for the classification of images based on the combination of NLP methods with image classification techniques. The goal was to classify objects not only according to a scheme of flat labels but in a more context-appropriate way, whereby the context informed by relevant domain-historical publications. This classification method offers access to the multidimensional embedding of artifacts in the real-world and their linguistic reflection. This circumstance is particularly useful for classifying multifunctional objects without having to resort to several classifiers and a complex annotation process with several labels. The results are encouraging. Even with a very simple  CNN, we achieve an accuracy of \num{0.54}.

As a next step, we want to train with a deeper CNN and an extended image corpus---to reduce known problems like overfitting. The comparative experiment with a conventional flat-label approach has shown that from an efficiency point-of-view, i.e., indirect metric comparison, our method not only provides comparable results but also provides a richer description of the image.

We will continue to work to better understand how a particular body of text is reflected int he labels that the DSM automatically assigns and that are not part of the list of target words. A better understanding of these processes seems particularly relevant given the relatively manageable text corpora that can be collated into specific topic complexes in the humanities. Last but not least, for this reason, we will consider the use of thesauri and dictionaries to create synonym lists for target words. Similarly, we are considering combining named entities into URIs. This would allow us to associate specific entities (e.g., workshops, ebenists, owners) with specific objects.

We think that multimodal access provides particularly efficient access to humanities and cultural studies corpora that are small and domain-restricted compared to corpora of other disciplines in the natural and social sciences.

\printbibliography

\end{document}


\maketitle

\section{Einleitend}
In den letzten Jahren hat die Verwendung von Bildklassifizierungsverfahren wie neuronalen Netzwerken auch im Bereich der historischen Bildwissenschaften und der \emph{Heritage Informatics} weite Verbreitung gefunden \parencite{lang_AttestingSimilaritySupportingOrganizationStudy_2018}. Diese Verfahren stehen dabei vor einer Reihe von Herausforderungen, darunter dem Umgang mit den vergleichsweise kleinen Datenmengen sowie zugleich hochdimensionalen Datenräumen in den digitalen Geisteswissenschaften. Meist bilden diese Methoden die Klassifizierung auf einen vergleichsweise flachen Raum ab. Dieser flache Zugang verliert im Bemühen um ontologische Eindeutigkeit eine Reihe von relevanten Dimensionen, darunter taxonomische, mereologische und assoziative Beziehungen zwischen den Klassen beziehungsweise dem nicht formalisierten Kontext. Eine in \parencite{donig_VomBildTextUndWieder_2019a} vorgeschlagene Lösung, diese Beziehungen wieder in den Prozess der Klassifizierung zurückzubringen, ist, sich die größere Ausdruckskraft von textbasierten Modellen zunutze zu machen, um die Fähigkeiten visueller Klassifikatoren zu erweitern. 

Dabei wird ein \mbox{\emph{Convolutional Neural Network} (CNN)} genutzt, dessen Ausgabe im Trainingsprozess, anders als herkömmlich, nicht auf einer Serie flacher Textlabel beruht, sondern auf einer Serie von Vektoren. Diese Vektoren resultieren aus einem \mbox{\emph{Distributional Semantic Model} (DSM)}, welches aus einem Domäne-Textkorpus generiert wird. Ein DSM ist ein multidimensionaler Vektorraum, in dem Wörter als Vektoren abgebildet werden \parencite{lenci_DistributionalModelsWordMeaning_2018a}.

Das durchgeführte Experiment beruht auf der Kollation von zwei Korpora: Einem textbasierten und einem visuellen. Mit dem Textkorpus wird zunächst ein DSM erzeugt und diesem dann eine Auswahlliste von Zielwörter zugeführt (die funktional den Annotationslabeln der Bilder entspricht). Als Ergebnis erhalten wir Vektoren, die mit diesen Wörtern korrespondieren und mit denen die Bilder annotiert werden. Mit diesen Vektorannotationen wird ein neuronales Netzwerk trainiert, das anschließend dem Klassifikator unbekanntes Bildmaterial identifizieren soll. Als Ergebnis dieses Klassifikationsprozesses erhalten wir einen Vektor, der mithilfe des DSMs in natürlichsprachige Wörter zurückgewandelt wird. Da wir nach reicheren Repräsentationen im Zuge dieses Vorgangs suchen, wählen wir die fünf nächsten Nachbarn aus (\mbox{\emph{Top-5 Nearest Neighbors}}). Als Ähnlichkeitsmaß legen wir die Kosinusähnlichkeit zwischen dem vorhergesagtem Vektor und jenem Vektor zugrunde, der dem ursprünglich dem Bild von uns zugewiesenen Textlabel entspricht. Wir gehen davon aus, dass ein Bild korrekt klassifiziert wurde, wenn das Goldlabel unter diesen Top-5 erscheint.

Darüber hinaus vergleichen wir die Ergebnisse des vorgeschlagenen Klassifizierungsverfahrens mit einem herkömmlichen Verfahren auf der Grundlage flacher Label unter Verwendung desselben CNNs, das für das Vektor-Experiment genutzt wurde. Wir können zeigen, dass das Vektor-Verfahren (bezogen auf die Metriken) ebenso effizient und in einigen Aspekten sogar besser ist.

\section{Aufbau des Experiments}
Das Experiment beruht auf je einem Bild- und Textkorpus aus dem Bereich Sachkulturforschung mit einem Fokus auf klassizistische Artefakte.

\subsection{Textkorpus}
Das Textkorpus besteht aus 44 Quellen, die unter einer freien, permissiven Lizenz verfügbar sind, und umfasst englischsprachige Fachpublikationen zu Mobiliar und Raumkunst, die von  der Jahrhundertwende bis zur Mitte des 20.~Jahrhunderts erschienen sind. Das Textkorpus wurde in mehreren Schritten gereinigt und aufbereitet: Zum einen wurden Standard-NLP-Verfahren angewandt, darunter Tokenisierung, Satz- und Worttrennung, die Normalisierung von Zahlenwerten und die Erkennung von benannten Entitäten (\emph{Named Entity Recognition, NER}). Da wir retrodigitalisiertes Material aus verschiedenen Quellen nutzen, implementieren wir manuelle Korrekturen für die häufigsten der vorkommenden Fehler (etwa Ligaturen wie \texttt{II}, die als \texttt{U} durch das OCR-Verfahren fehlinterpretiert werden). Eine weitere Ebene der Vorverarbeitung besteht aus inhaltsbezogenen Augmentierungen. Insbesondere normalisieren wir zusammengesetzte Wörter und Synonyme gemäß einer spezifizierten Liste, die anhand einer Ontologie, der \emph{Neoclassica-Ontologie} \parencite{donig_NeoclassicaMultilingualDomainOntology_2016} zusammengestellt sind. Dies resultiert in einem Korpus von insgesamt \num{3067237} Wörtern aus \num{107518} Wortgrundformen.
Das DSM wird von uns mithilfe des \emph{Indra}-Frameworks \parencite{sales_IndraWordEmbeddingSemanticRelatedness_2018a} sowie \emph{Gensim} \parencite{rehurek_lrec} erzeugt\footnote{Die Erzeugung erfolgt mit einer Vektorgröße von 50, einer Worfenstergröße von 10 und einer minimalen Wortzahl von fünf erzeugt. Als \emph{Word2Vec}-Modell kommt \emph{Skipgram} \parencite{mikolov_EfficientEstimationWordRepresentationsVector_2013} mit \emph{Negative Sampling} zum Einsatz.}.

\subsection{Bildkorpus}
Das Bildkorpus besteht aus \num{1231} Ansichten klassizistischer Möbel in deren Gesamtheit, die permissiv lizenziert sind\footnote{Das Korpus wurde aus den Sammlungen des Metropolitan~Museum (New~York), des Victoria~\&~Albert~Museum (London), der Wallace~Collection (London) sowie mehreren zeitgenössischen Musterbüchern zusammengestellt.} und die sowohl historisches Bildmaterial als auch Fotos aus der modernen Bestandsdokumentation umfassen. Es repräsentiert 28 Klassen.

\subsection{Kombiniertes Korpus}
Da es sich um ein \emph{Proof-of-Concept}-Experiment handelt, kommt zum Zweck des \emph{Rapid Prototypings} ein an die VGG-Architektur \parencite{krizhevsky_ImageNetClassificationDeepConvolutionalNeural_2012a} angelehntes, \enquote{simples} neuronales Netzwerk zum Einsatz\footnote{Das Netzwerk besteht aus drei Convolutional-Blöcken mit jeweils zwei Convolutional-Layer mit \numlist{32;64;64} Filter der Größe $ 3 \times 3 $. Nach jedem Block folgt ein Maximum-Pooling-Layer der Größe $ 2 \times 2 $, sowie ein Dropout-Layer mit einer Dropoutwahrscheinlichkeit von \num{0.25}. Ein Fully-Connected-Block, bestehend aus zwei Fully-Connected-Layers mit jeweils 256 Knoten, steht im Anschluss sowie nochmals ein Dropout-Layer mit \num{0.5} Dropoutwahrscheinlichkeit. Jeder Convolutional- und Fully-Connected-Layer bis dahin wurde zufällig initialisiert und benutzt \emph{ReLU} als Aktivierungsfunktion. Der letzte Layer ist ein Fully-Connected-Layer mit 50 Ausgabeknoten und benutzt eine lineare Aktivierungsfunktion. Es ist in den beiden Frameworks \emph{Keras} (\url{https://keras.io}) und \emph{TensorFlow} (\url{https://tensorflow.org}) implementiert. Beim Training wird der durchschnittliche absolute Fehler durch die Optimierungsfunktion \emph{RMSprop} minimiert.}.
Die Unabhängigkeit der Trainings- und Evalutationsbeispiele wird durch eine \mbox{5-fache} Kreuzvalidierung auf \SI{80}{\percent} des Korpus garantiert, wovon wiederum \SI{80}{\percent} für das Training und \SI{20}{\percent} für das Testen benutzt werden. Die übrigen \SI{20}{\percent} werden als Evaluationsmenge behandelt. Die reportierten Ergebnisse basieren auf diesen, dem Klassifikator unbekannten \SI{20}{\percent}. 

\begin{figure}
	\centering
	\includegraphics[height=.5\textheight]{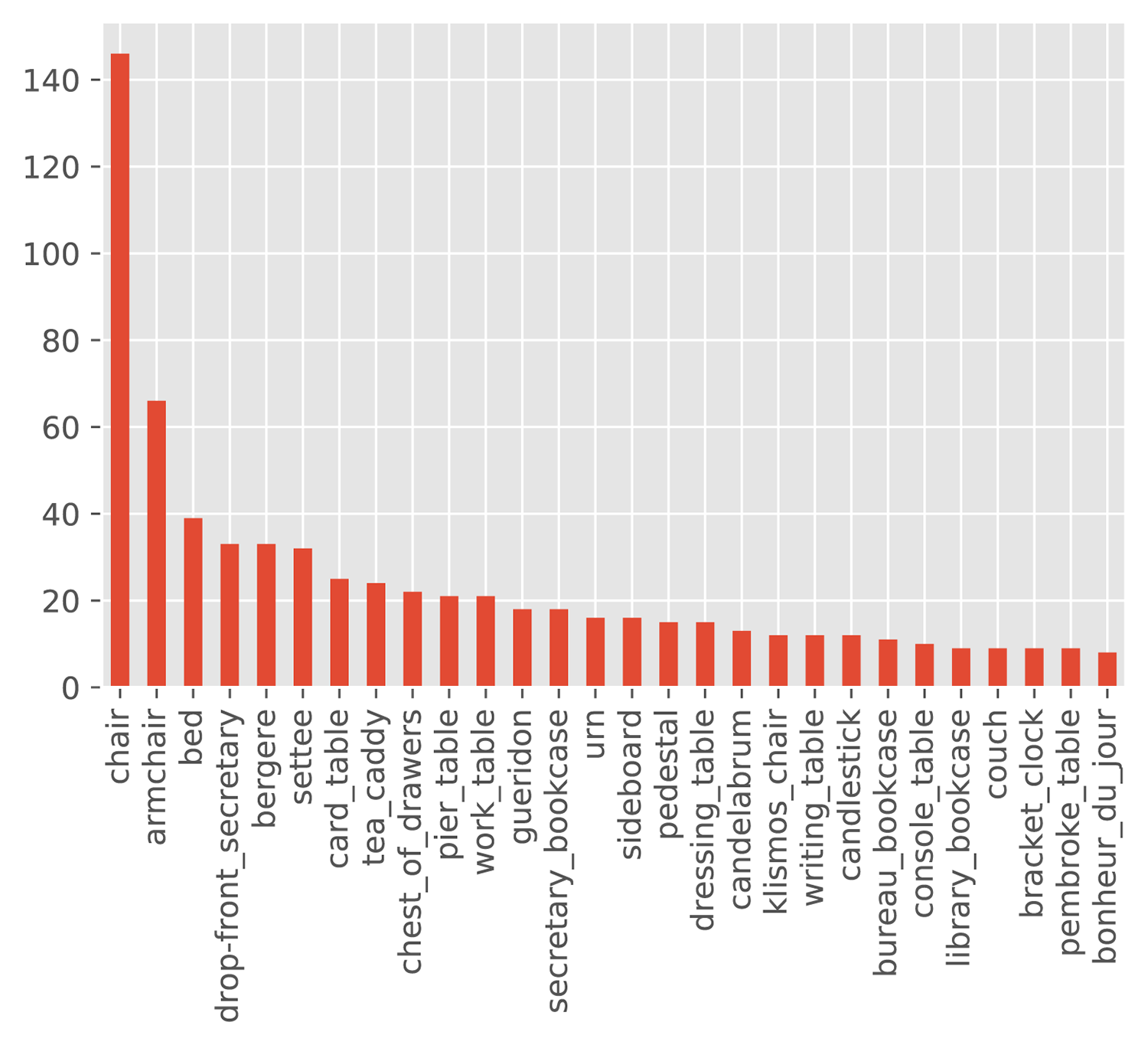}
	\caption{Verteilung der Klassen im Bildkorpus.}
	\label{fig:class_distribution}
\end{figure}

Da durch Sammlungspraxis der Gedächtnisinstitutionen (Sammelwürdigkeit, geografischer Schwerpunkt) und Zugänglichkeit des Materials (Lizenzierung und Grad der Sammlungsdigitalisierung) die Verteilung der Artefakte nach Klassen unbalanciert ist (Abb.~\ref{fig:class_distribution}), sind die Klassengewichte dementsprechend angepasst (seltene Klassen werden höher gewichtet als häufig vorkommende Klassen \parencite[S.~27]{johnson_SurveyDeepLearningClassImbalance_2019}). Um eine Situation zu vermeiden, in der ein \emph{Machine-Learning}-Modell derart an die Trainingsdaten angepasst wird, dass es darin scheitert auf anderen, aber trotzdem ähnlichen Daten, zu generalisieren, wird die übliche \emph{Early-Stopping}-Methode verwendet.

\section{Ergebnisse}
Die \mbox{Top-5-Richtig-Positiv-Rate} (\emph{true-positive}) beträgt \num{0.59}. Das bedeutet, dass das Goldlabel in \SI{59}{\percent} der Fälle unter den fünf nächsten Nachbarn erscheint.

Das mathematische Qualitätskriterium gibt für sich genommen jedoch nur einen Teil des Gesamtbilds wieder. Wir führen deshalb zugleich eine qualitative Analyse der Ergebnisse in der Evalutionsteilmenge durch.

\begin{figure}
	\centering
	\includegraphics[height=.4\textheight]{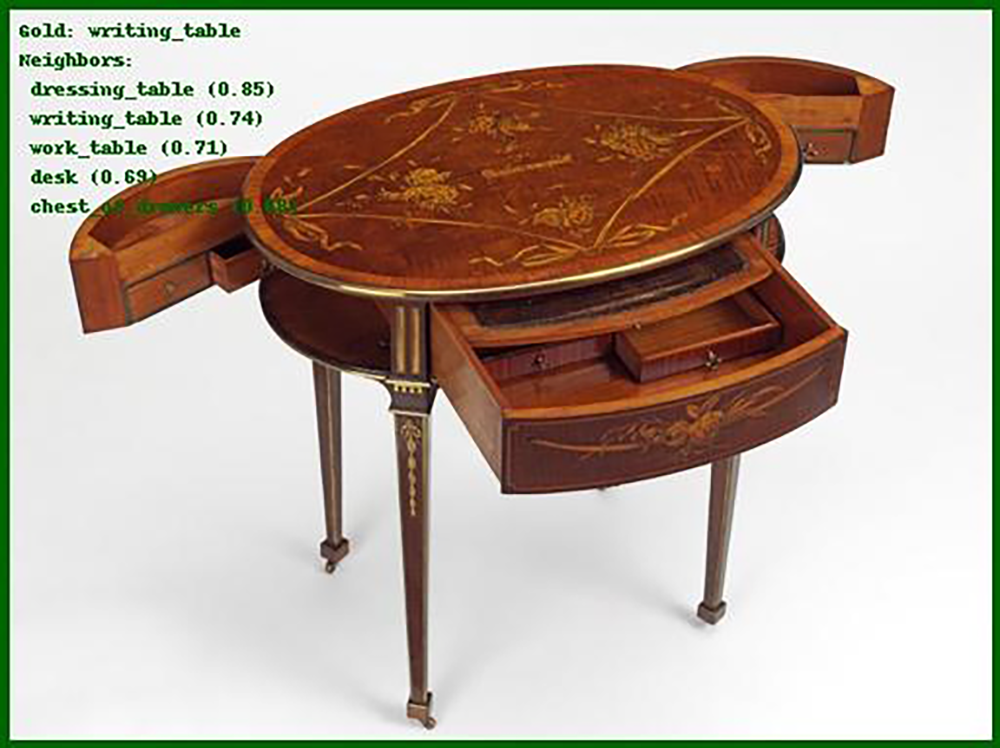}
	\caption{Abweichungen bei der Klassifizierung desselben Objekts \parencite{victoriaalbertmuseum_WritingTableNeuwiedWorkshopDavid_2019}.}
	\label{fig:classification_same_object_1}
\end{figure}

\begin{figure}
	\centering
	\includegraphics[height=.4\textheight]{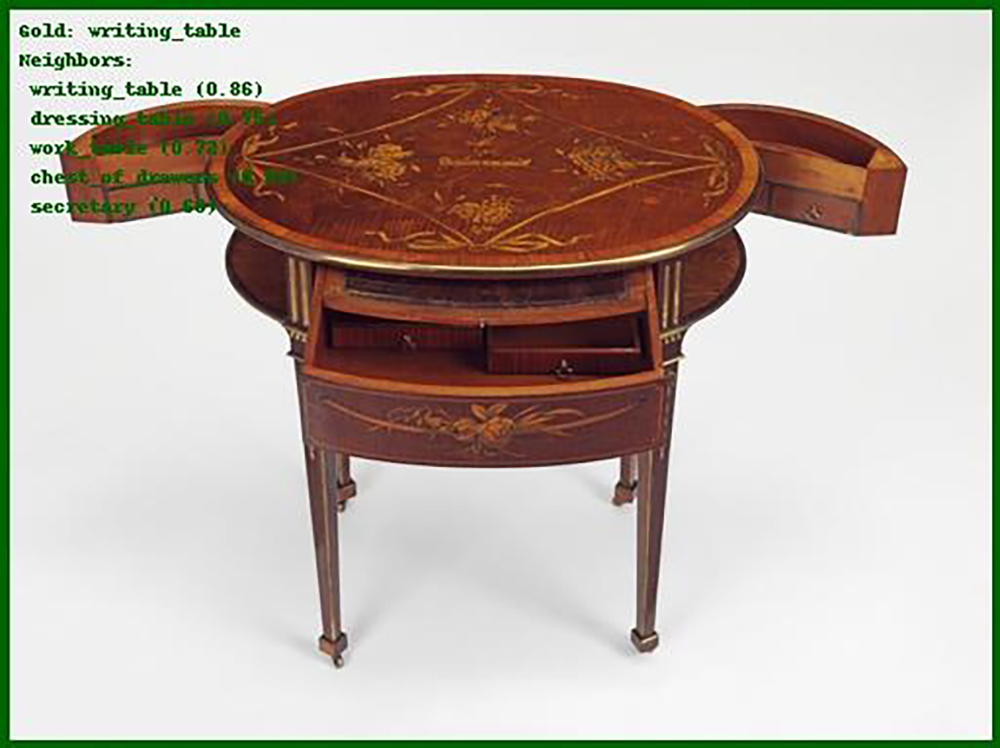}
	\caption{Abweichungen bei der Klassifizierung desselben Objekts \parencite{victoriaalbertmuseum_WritingTableNeuwiedWorkshopDavid_2019}.}
	\label{fig:classification_same_object_2}
\end{figure}

Eine Reihe von richtig-positiven Ergebnissen zeigen, dass die Klassifizierung keinesfalls zufällig erfolgt, sondern dass die \mbox{Top-5-Begriffe} tatsächlich jeweils denselben semantischen Nachbarschaften entstammen. Sie drücken eine Reihe von Beziehungen taxonomischer und assoziativer Natur aus.
Beispielsweise wird der \mbox{Roentgen-Schreibtisch} aus dem Bestand des V\&A in Abb.~\ref{fig:classification_same_object_1} mit Labeln (in der Reihenfolge) von \texttt{dressing\_table}, \texttt{writing\_table} und \texttt{work\_table} assoziiert (Ankleidetisch, Schreibtisch, Nähtisch). Diese Trias ist schon deshalb sinnhaltig, weil viele dieser Artefakte multifunktional waren und mehrere dieser Funktionen erfüllten. Daneben ähneln auch jene Artefakte, die dezidiert nur einem einzigen Zweck dienten, konstruktiv den jeweils anderen Möbeltypen. Die Nähe der drei Konzepte entsteht also sowohl auf semantischer Ebene (Nähe der Wörter im DSM, die wiederum das Produkt lebensweltlicher Nähe ist), als auch auf einer visuellen Ebene im CNN (visuelle Formähnlichkeit). Ein weiteres Bild desselben Objekts (Abb.~\ref{fig:classification_same_object_2}) zeigt einerseits, dass die Methode in sich konsistent ist (die Top-4 sind identisch, obwohl eine andere Perspektive vorliegt) und andererseits, dass auch die visuellen Merkmale innerhalb des CNNs eine Auswirkung auf den Klassifizierungsprozess haben. Da Schreibschränke (\emph{secrétaires à abbatants}) häufig frontal, hochaufrecht und mit einer geöffneten Schreibklappe oder -schublade abgebildet werden, scheint deren Vorkommen im Bild eine Klassifizierung als Sekretär ausgelöst zu haben. Im ersten Bild könnte dagegen die Anwesenheit von Schubladen (\emph{drawers}) zu einer Klassifizierung als Kommode geführt haben, die naheliegenderweise auf semantischer Ebene mit Schubladen assoziiert ist.

\begin{figure}
	\centering
	\includegraphics[height=.6\textheight]{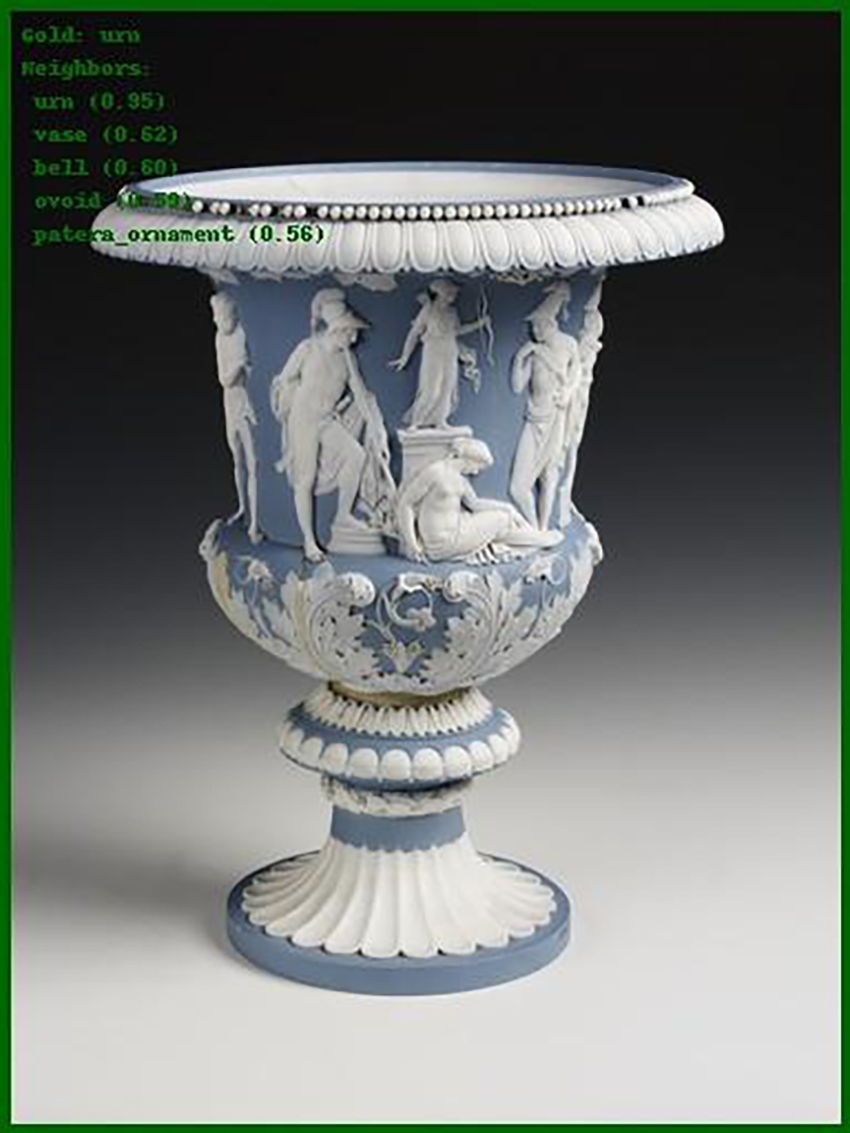}
	\caption{Eine Sèvres-Kopie der Medici-Vase stößt die Klassifizierung mit assoziativen Labeln an. \parencite{victoriaalbertmuseum_VaseSevresCopyMediciVase_2019}}
	\label{fig:medici_vase}
\end{figure}

Während die Label in den bisher betrachteten Fällen die taxonomischen Beziehungen reflektieren und alle den aus der Ontologie abgeleiteten Zielwörtern entstammen, zeigt Abb.~\ref{fig:medici_vase}, dass das Verfahren auch aus sich selbst, rein datenzentriert Label generieren kann. Die abgebildete Kratervase wurde manuell als Urne (\emph{urn}) annotiert. Die \mbox{Top-2-Wörter} reflektieren demnach auch taxonomischen Beziehungen (\mbox{\enquote{urn}}, \mbox{\enquote{vase}}). Die anderen Konzepte spiegeln dagegen assoziative Beziehungen wider. Das Label \texttt{bell} ist ein Artefakt des Reinigungsprozesses, da im Korpus Wörter wie \mbox{\enquote{bell-shaped}} oder \mbox{\enquote{bell-crater}} (mit und ohne Bindestrich) existieren, um diese Art von Artefakten zu beschreiben. \enquote{Ovoid} bezieht sich demgegenüber wohl auf die Eierstabdekoration des oberen Wulstes, die oft mit diesem Adjektiv beschrieben wird. Diese Ornamentik scheint zugleich die Assoziation zur Rosette (\emph{Patera}) mitbedingt zu haben. Auf diese Weise erscheint das Zielwort \texttt{patera\_element} unter den \mbox{Top-5}, obwohl im Bildkorpus ausschließlich ganze Artefakte, nicht aber deren Dekor annotiert wurden. 

\begin{figure}
	\centering
	\includegraphics[height=.6\textheight]{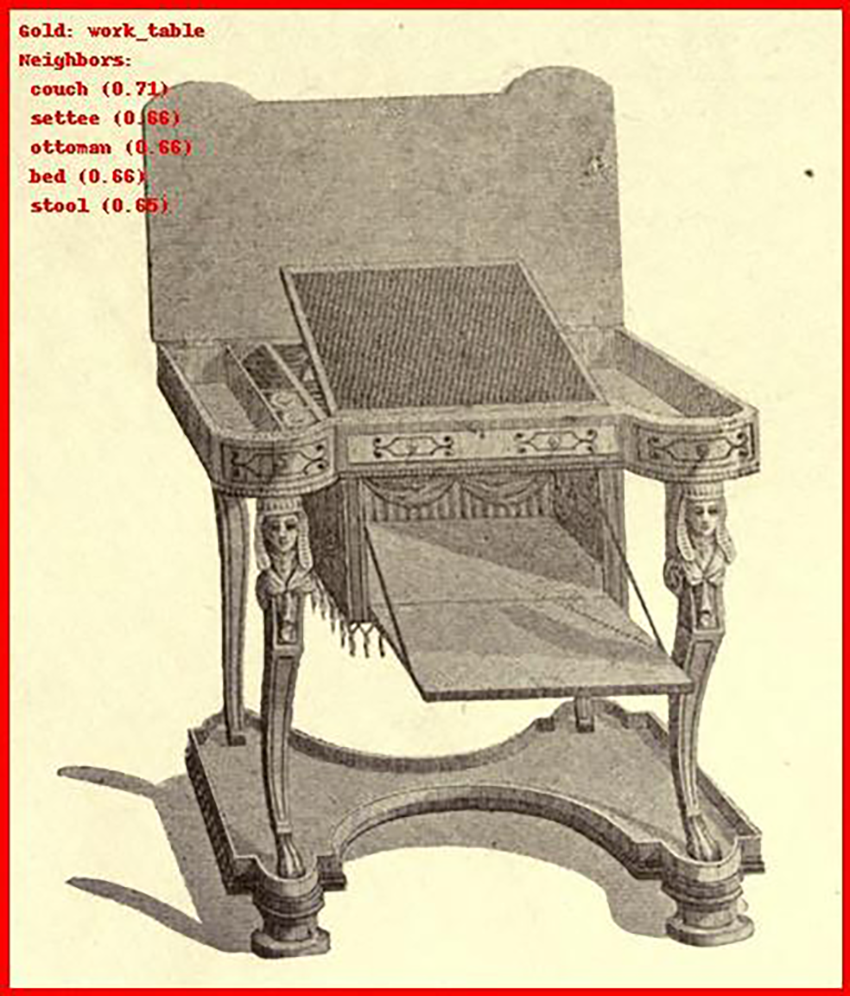}
	\caption{Fehlklassifikation eines Nähtischs in ein Wortumfeld aus der Sitzmöbel-Hierarchie. \parencite[S.~88]{sheraton_FurnitureDesignsThomasSheraton_1910}}
	\label{fig:misclassification}
\end{figure}

Nicht auszuschließen ist hier zudem ein Effekt des visuellen Klassifikators, wie auch Abb.~\ref{fig:misclassification} zeigt. Die Fehlklassifikation des Objekts, eines Nähtischchens, führte zu konsistenten Zuschreibungen im Bereich der Sitz- und Liegemöbel. Betrachtet man die äußere Form des Artefakts auf einer abstrakteren Ebene, kann man eine visuelle Nähe zu z.~B.~einem \emph{(Double-) Camel-back Sofa} durchaus nachvollziehen.

\section{Vergleich zu einem herkömmlichen CNN}

Um die Unterschiede zwischen beiden Zugängen besser abschätzen zu können, führen wir weiter in Vergleichsexperiment durch, bei dem dasselbe CNN\footnote{Das verwendete CNN ist identisch mit dem im Vektor-Experiment verwendeten Netzwerk, mit Ausnahme des letzten Layers, der aufgrund der Labelklassifikation nun 28 statt 50 Ausgaben besitzt und die Aktivierungsfunktion \emph{Softmax} statt einer linearen.} wie im vektorbasierten Verfahren für eine herkömmliche Klassifizierung mit flachen Labeln herangezogen wird.

Um die Unterschiede der beiden Zugänge (herkömmliche Klassifizierung mit flachen Labeln/vektorbasiertes Verfahren) besser abschätzen zu können, stellen wir im Folgenden einen Vergleich zentraler Metriken für beide Zugänge vor.

\begin{table}
    \centering
    \begin{tabular}{lrr}
    	\toprule
    	Metrik                    &     Vektor & Flache Label \\ \midrule
    	Top-1-Treffergenauigkeit  & \num{0.50} &   \num{0.40} \\
    	Top-1-Genauigkeit         & \num{0.32} &   \num{0.29} \\
    	Top-1-Trefferquote        & \num{0.25} &   \num{0.26} \\
    	Top-1-F1-Maß              & \num{0.26} &   \num{0.25} \\
    	Top-5-Sensitivität        & \num{0.59} &   \num{0.73} \\
    	Top-5-Falsch-Positiv-Rate & \num{0.41} &   \num{0.27} \\
         \bottomrule
    \end{tabular}
    \caption{Vergleich zentraler Metriken beider Zugänge.}
    \label{tab:results}
\end{table}

Wie in Tabelle~\ref{tab:results} ersichtlich\footnote{Die englischen Begriffe für bessere Verständlichkeit: Treffergenauigkeit (\emph{Accuracy}), Genauigkeit (\emph{Precision}), Trefferquote (\emph{Recall}), Sensitivität (\emph{True-Positive Rate})}, ist nicht nur die \mbox{Top-1-Treffergenauigkeit} im Fall der Klassifizierung mit Vektoren besser, sondern es sind auch die übrigen Metriken vergleichbar gut. Durch den hier vorgeschlagenen Zugang wird also nicht nur die Treffergenauigkeit verbessert, sondern er liefert zugleich eine reichhaltigere Beschreibung des Bildes.

\section{Schlussfolgerungen und Ausblick}
In dem hier vorgeschlagenen Beitrag stellen wir ein neues, multimodales Verfahren für die Klassifikation von Bildinhalten vor, das auf der Kombination von NLP-Methoden mit Bildklassifizierungsverfahren beruht. Ziel ist, Objekte nicht alleine nach einem Schema flacher Label, sondern in einer kontextgerechteren Weise zu klassifizieren, wobei dieser Kontext von einschlägigen historischen Domänepublikationen gebildet wird. Dieses Klassifizierungsverfahren bietet einen Zugang zur multidimensionalen Einbettung der Artefakte in die Lebenswelt und deren sprachlicher Widerspiegelung. Dieser Umstand ist von besonderem Nutzen um multifunktionale Objekte zu klassifizieren, ohne dabei auf mehrere Klassifikatoren und einen komplexen Annotationsprozess mit mehreren Labeln zurückgreifen zu müssen. Die Ergebnisse sind ermutigend. Auch mit einem sehr einfachen CNN erreichen wir eine Treffergenauigkeit von \num{0.59}. Als nächsten Schritt möchten wir mit einem komplexeren CNN und einem ausgeweiteten Bildkorpus trainieren (um bekannte Probleme wie \emph{Overfitting} zu reduzieren). Das Vergleichsexperiment mit einem herkömmlichen, auf flachen Labeln beruhenden Zugang zeigt, dass unter Effizienzgesichtspunkten, d.~h.~im direkten Vergleich der Metriken, unser Verfahren nicht nur vergleichbare Resultate liefert, sondern zugleich auch in einer reichhaltigeren Beschreibung des Bildes resultiert.

Wir werden weiter daran arbeiten, besser zu verstehen, wie ein bestimmtes Textkorpus sich in den Labeln widerspiegelt, die das DSM automatisch zuweist und die nicht Teil der Liste der Zielwörter sind. Ein besseres Verständnis dieser Prozesse scheint insbesondere im Hinblick auf die relativ überschaubaren Textkorpora relevant, die in den Geisteswissenschaften zu spezifischen Themenkomplexen kollationiert werden können. Nicht zuletzt werden wir aus diesem Grund die Nutzung von Thesauri und Wörterbüchern in Betracht ziehen, um Synonymlisten für Zielwörter zu erstellen. In ähnlicher Weise ziehen wir in Betracht, benannte Entitäten zu URIs zusammenzufassen. Das würde uns erlauben spezifische Entitäten (z.~B.~Werkstätten, Ebenisten und Eigentümer) mit bestimmten Objekten zu assoziieren.

Wir denken, dass dabei der multimodale Zugriff einen besonders effizienten Zugang zu geistes- und kulturwissenschaftlichen Korpora bietet, die, verglichen mit den Korpora anderer Disziplinen in den Natur- und Sozialwissenschaften, klein und Domäne-restringiert sind.

\printbibliography